\pdfoutput=1
\documentclass[english]{llncs}
\usepackage[T1]{fontenc}
\usepackage[latin9]{inputenc}
\usepackage{amssymb}
\usepackage{babel}
\usepackage{stackrel}
\usepackage{epsfig}
\usepackage{svg}
\usepackage{amsmath}
\usepackage{graphicx}

\usepackage{epstopdf}
\usepackage{booktabs}

\def\v#1{\mathbf{#1}}

\begin{document}
\title{Higher Order of Motion Magnification for Vessel Localisation in Surgical Video}
\titlerunning{Higher Order Motion Magnification}  
%
\author{Mirek Janatka$^1$ \and Ashwin Sridhar\and John Kelly  \and Danail Stoyanov$^1$}
\authorrunning{Mirek Janatka et al.} 
%
\tocauthor{Mirek Janatka, Ashwin Sridhar, John Kelly, Danail Stoyanov}
\institute{Wellcome / EPSRC Centre for Interventional and Surgical Sciences, University College London, United Kingdom\\
$^1$Department of Computer Science, University College London, United Kingdom
\email{mirek.janatka@ucl.ac.uk}}
\maketitle         
\begin{abstract}
Locating vessels during surgery is critical for avoiding inadvertent damage, yet vasculature can be difficult to identify. Video motion magnification can potentially highlight vessels by exaggerating subtle motion embedded within the video to become perceivable to the surgeon. In this paper, we explore a physiological model of artery distension to extend motion magnification to incorporate higher orders of motion, leveraging the difference in acceleration over time (jerk) in pulsatile motion to highlight the vascular pulse wave. Our method is compared to first and second order motion based Eulerian video magnification algorithms. Using data from a surgical video retrieved during a robotic prostatectomy, we show that our method can accentuate cardio-physiological features and produce a more succinct and clearer video for motion magnification, with more similarities in areas without motion to the source video at large magnifications. We validate the approach with a Structure Similarity (SSIM) and Peak Signal to Noise Ratio (PSNR) assessment of three videos at an increasing working distance, using three different levels of optical magnification. Spatio-temporal cross sections are presented to show the effectiveness of our proposal and video samples are provided to demonstrates qualitatively our results. 
\end{abstract}
\section{Introduction}
\begin{figure}[!ht]
\centering
\includegraphics[width=\textwidth]{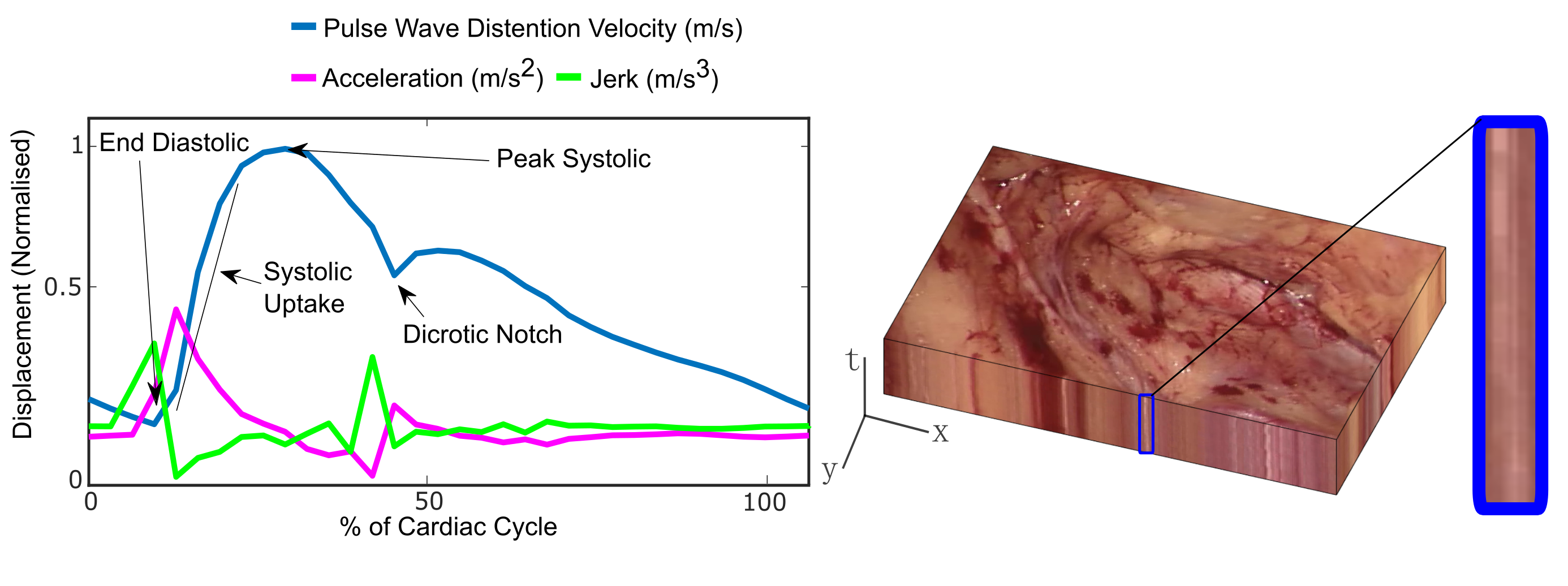}
\caption{(Left): The vessel distension-displacement from the pulse wave, with the higher order derivatives along with annotation of the corresponding cardio-physiological stages. Down sampled to 30 data points to reflect endoscope frame rate acquisition. (1-D Virtual Model of arterial behaviour \cite{VirtualPulseWave}) (Right): Endoscopic video image stack. The blue box surrounds an artery with no perceivable motion, shown by the vertical white line in the cross section}
\label{fig:OneDimension}
\end{figure}

One of the most common surgical complications is due to inadvertent damage to blood vessels. Avoiding vascular structures is particularly challenging in minimally invasive surgery (MIS) and robotic MIS (RMIS) where the tactile senses are inhibited and cannot be used to detect pulsatile motion. Vessels can be detected by using interventional imaging modalities like fluorescence or ultrasound (US) but these do not always produce a sufficient signal, or are difficult to use in practice \cite{AshwinAR}. Using video information directly is appealing because it is inherently available, but processing is required to reveal any vessel information hidden within the video and is not apparent to the surgeon, as can be seen in the right image of Fig.\ref{fig:OneDimension}.

The cardiovascular system creates a pressure wave that propagates through the entire body and causes an equivalent distension-displacement profile in the arteries and veins \cite{PulseWaveHaemo}. This periodic motion has intricate characteristics, shown in Fig.~\ref{fig:OneDimension} (left), that can be highlighted by differentiating the distension-displacement signal. The second order derivative outlines where the systolic uptake is located, whilst the third derivative highlights the end diastolic phase and the dicrotic notch. This information can be present as spatio-temporal variation between image frames and amplified using Eulerian video magnification (EVM). EVM could be applied to endoscopic video for vessel localisation by using an adaptation of an EVM algorithm and showing the output video directly to the surgeon \cite{McLeod2014}. Similarly, EVM can aid vessel segmentation for registration and overlay of pre-operative data \cite{KhaliliAutoSeg}, as existing linear based forms of the raw magnified video can be abstract and noisy to use directly within a dynamic scene. Magnifying the underlying video motion can exacerbate unwanted artifacts and unsought motions, and in this case regarding surgical video, of those which are not the blood vessels but due to respiration, endoscope motion or other physiological movement within the scene.

In this paper, we propose to utilise features that are apparent in the cardiac pulse wave, particularly the non-linear motion components that are emphasised by the third order of displacement, known as jerk (Green plot Fig.~\ref{fig:OneDimension}, left). We devise a custom temporal filter and use an existing technique for spatial decomposition of complex steerable pyramids \cite{Simoncelli1990ChapterTransforms}. The result is a more coherent magnified video compared to existing lower order of motion approaches \cite{Wadhwa2012,ZhangAccelerationMag}, as the high magnitudes of jerk are prominently exclusive to the pulse wave in the surgical scene, as our method avoids amplification of residual motions due to respiration or other periodic scene activities. Quantitative results are difficult for such approaches but we report a comparison to previous work using Structure Similarity \cite{SSIM} and Peak Signal to Noise Ratio (PSNR) of three robotic assisted surgical videos at separate optical zoom. We provide a qualitative example of how our method achieves isolation of two cardio-physiological features over existing methods. A supplementary video of the magnifications is provided that further illustrates the results.
\section{Methods}
Building on previous work in video motion magnification \cite{ZhangAccelerationMag} \cite{Wadhwa2012} \cite{EVMSource} we set out to highlight the third order motion characteristics created by the cardiac cycle. In an Eulerian frame of reference, the input image signal function is taken as $I(\v x,t)$ at position $\v x$ ($\v x = (x,y)$)  and at time \textit{t} \cite{EVMSource}. With the linear magnification methods, $\delta(t)$ is taken as a displacement function with respect to time, giving the expression $I(\v x,t) = f(\v x + \delta(t))$ and is equivalent to the first-order term in the Taylor expansion:
\begin{equation}
I(\v x,t) \thickapprox f(\v x)+\delta(t)\frac{\partial f(\v x)}{\partial \v x}
\end{equation}
This Taylor series expansion appropriation can be continued into higher orders of motion, as shown in \cite{ZhangAccelerationMag}. Taking it to the third order, where $\hat{I}(\v x,t)$  is the magnified pixel at point $\v x$ and time $t$ in the video.
\begin{equation}
\hat{I}(\v x,t) \thickapprox f(\v x)+(1+\beta)\delta(t)\frac{\delta f(\v x)}{\delta \v x}
+(1+\beta)^{2}\delta(t)^{2}\frac{1}{2}\frac{\delta^{2}f(\v x)}{\delta^{2}\v x}
+(1+\beta)^{3}\delta(t)^{3}\frac{1}{6}\frac{\delta^{3}f(\v x)}{\delta^{3}}\\
\end{equation}
In a similar vein to \cite{ZhangAccelerationMag}, we equate a component of the expansion to an order of motion and isolate these by subtraction of the lower orders
\begin{equation}
I(\v x,t) - I(\v x,t)_{non-linear(2^{nd}order)} - I(x,t)_{linear}\thickapprox
(1+\beta)^{3}\delta(t)^{3}\frac{1}{6}\frac{\delta^{3}f(x)}{\delta^{3}\v x}
\end{equation}
assuming (1+$\beta)^{3}$= $\alpha,$
$\alpha>0$. 
\begin{equation}
D(\v x,t) = \delta(t)^{3}\frac{1}{6}\frac{\delta^{3}f(\v x)}{\delta^{3}\v x}
\end{equation}
\begin{equation}
\hat{I}_{non-linear(3^{nd}order)}(\v x,t) = I(\v x,t) + \alpha D(\v x,t)
\end{equation}
This produces an approximation for for the input signal and a term that can be attenuated in order to present an augmented reality (AR) view of the original video.
\subsection{Temporal Filtering}
As jerk is the third temporal derivative of the signal $\hat{I}(\v x,t)$, a filter has to be derived to reflect this. To achieve acceleration magnification, the Difference of Gaussian (DoG) filter was used \cite{ZhangAccelerationMag}. This allowed for a temporal bandpass to be assigned, by subtracting two Gaussian filters, using $\sigma$ = $\frac{r}{4\omega\sqrt{2}}$ \cite{Mikolajczyk2001IndexingPoints} to calculate the standard deviations of them both, where \textit{r} is the frame rate of the video and $\omega$ is the frequency under investigation. Taking the derivative of the second order DoG we create an approximation of the third order, which follows Hermitian polynominals \cite{2003GaussianDerivatives}. Due to the linearity of the operators, the relationship between the the jerk in the signal and the third order DoG as:
\begin{equation}
\frac{\partial^{3}I(\v x,t)}{\partial t^{3}}\otimes G_{\sigma}(t)
= I(\v x,t) \otimes\frac{\partial^{3}G_{\sigma}(t)}{\partial t^{3}}
\end{equation}
\subsection{Phase-based Magnification}
In the classical EVM approach, the intensity change over time is used in a pixel-wise manner \cite{EVMSource} where a second order IIR filter detects the intensity change caused by the human pulse. An extension of this uses the difference in phase w.r.t spatial frequency \cite{Wadhwa2012} for linear motion, as subtle difference in phase can be detected between frames where minute motion is present. Recently, phase-based acceleration magnification has been proposed \cite{ZhangAccelerationMag}. It is this methodology we utilise and amend for jerk magnification. By describing motion as phase shift, a decomposition of the signal $f(x)$ with displacement $\delta(t)$ at time t, the sum of all frequencies ($\omega$) can be shown as:
\begin{equation}
f(\v x+\delta(t))=\stackrel[\omega=-\infty]{\infty}{\sum}A_{\omega}e^{i\omega(\v x+\delta(t))}
\end{equation}
where the global phase for frequency $\omega$ for displacement $\delta(t)$ is $\phi_{\omega} = \omega(\v x + \delta(t))$.

It has been shown that spatially localised phase information of a series of image over time is related to local motion \cite{Portilla2000ACoefficients} and has been leveraged for linear magnification \cite{Wadhwa2012}. This is performed by using complex steerable pyramids \cite{portilla2000parametric} to separate the image signal into multi-frequency bands and orientations. These pyramids contain a set of filters $\Psi_{\omega_{s},\theta}$ at multiple scales, $\omega_{s}$ and orientations $\theta$. The local phase information of a single 2D image $I(\v x)$ is
\begin{equation}
(I(\v x))\otimes\Psi_{\omega_{s},\theta}(\v x) = A_{\omega,\theta}(\v x)e^{i\phi_{\omega_{s},\theta}(\v x)}
\end{equation}
Where $A_{\omega,\theta}(\v x)$ is the amplitude at frequency $\omega$ and orientation $\theta$, and where $\phi_{\omega_{s},\theta}$ is the corresponding phase at scale (pyramid level) $\omega_{s}$.   
The phase information is extracted ($\phi_{\omega_{s},\theta}(\v x,t)$) at a given frequency $\omega$, orientation $\theta$ and frame \textit{t}. The jerk constituent part of the motion is filtered out with our third order Gaussian filter and can then be magnified and reinstated into the video ($\hat{\phi}_{\omega,\theta}(\v x,t)$) to accentuate the desired state changes in the cardiac cycle, such as the dicrotic notch and end diastolic point, shown in Fig.~\ref{fig:OneDimension} (left).
\begin{equation}
D_{\sigma}(\phi_{\omega,\theta}(\v x,t))
= \phi_{\omega,\theta}(\v x,t)\otimes\frac{\partial^{3}G_{\sigma}(t)}{\partial t^{3}}
\end{equation}
\begin{equation}
\hat{\phi}_{\omega,\theta}(\v x,t) = \phi_{\omega,\theta}(\v x,t)
+\alpha D_{\sigma}\phi_{\omega,\theta}(\v x,t)
\end{equation}
Phase unwrapping is applied as with the acceleration methodology in order to create the full composite signal \cite{ZhangAccelerationMag}\cite{Kitahara2015AlgebraicStabilizations}.
\section{Results}
To demonstrate the proposed approach, endoscopic video was captured from robotic prostatectomy using the da Vinci surgical system (Intuitive Surgical Inc, CA), where a partially occluded obturator artery could be seen. Despite being identified by the surgical team the vessel produced little perceivable motion in the video. This footage was captured at 1080p resolution at 30Hz. For processing ease, the video was cropped to a third of the original width, which contained the motion of interest, yet still retains the spatial resolution of the endoscope. The video was motion magnified using the phase-based complex steerable pyramid technique described in \cite{Wadhwa2012} for first order motion and the video acceleration magnification described in \cite{ZhangAccelerationMag} offline for comparison. 
Our method appended the video acceleration magnification method. All processes use a four level pyramid and half octave pyramid type. For the temporal processing, a bandpass was set at 1Hz +/- 0.1 to account for a pulse around 54 to 66 bpm. From the patient's ECG reading, their pulse was stable at 60 bpm during video acquisition. This was done at three magnification factors (x2, x5, x10). Spatio-temporal slices were then taken of a site along the obturator artery for visual comparison of each temporal filter type. For a quantitative comparison, the Peak Noise to Signal Ratio (PNSR) and Structural Similarity (SSIM) index \cite{SSIM} was calculated on a hundred frame sample, comparing the magnified videos to their original equivalent frame.
\begin{figure}[!ht]
\centering
\includegraphics[width=0.9\textwidth]{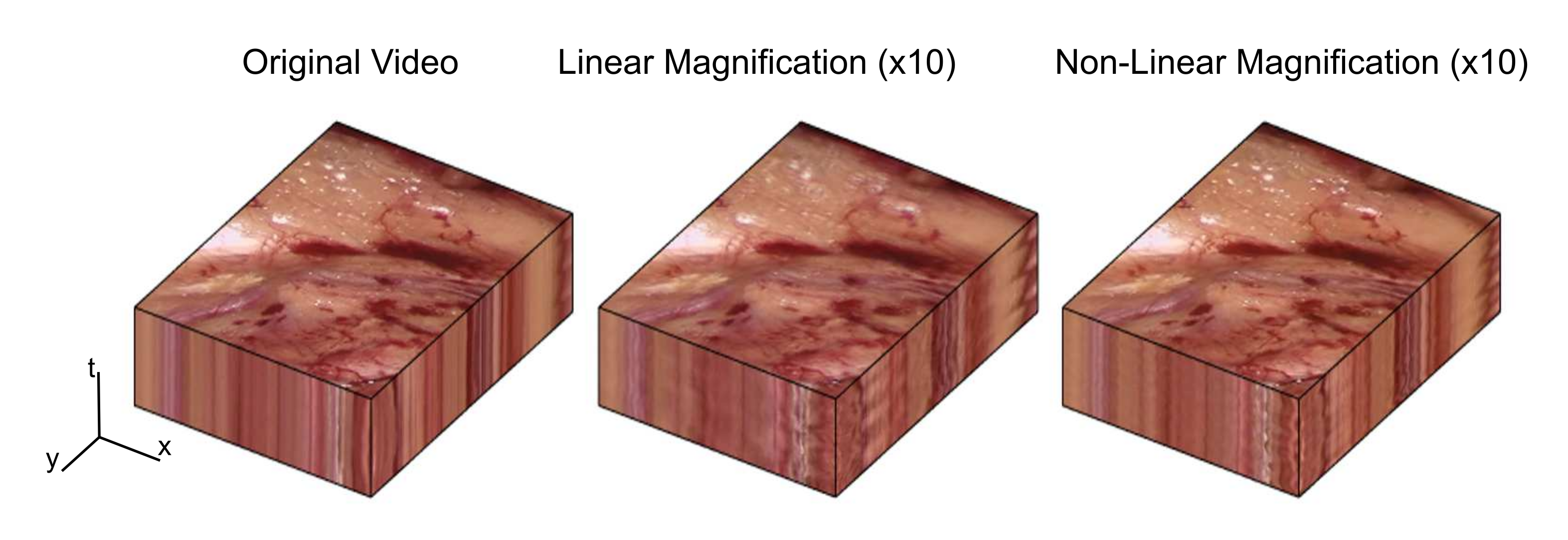}
\caption {Volumetric image stacks of an endoscopic scene under different types of magnification.}
\label{fig:MagCompareBox}
\end{figure}
\begin{figure}[!ht]
\centering
\includegraphics[width=0.95\textwidth]{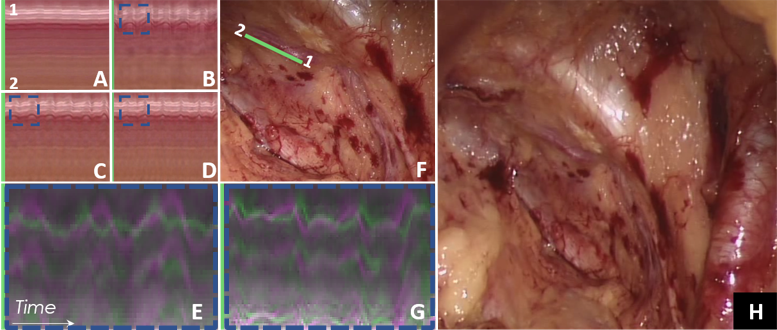}
\caption{ Motion Magnification of the obturator artery (x10). (a) Unmagnified spatio-temporal slice (STS); (b) Linear magnification \cite{Wadhwa2012}; (c) Acceleration magnification \cite{ZhangAccelerationMag}; (d) Jerk Magnification (our proposal); (e),(g) Comparative STS, blue box from (d) (jerk) in green, with (b) in magenta in (e) and (c) in magenta in (g); (f) Sample site (zoomed); (h) Overview of the surgical scene. }
\label{SurgicalSceneComparation}
\end{figure} 
\begin{figure}[!ht]
\centering
\includegraphics[width=0.7\textwidth]{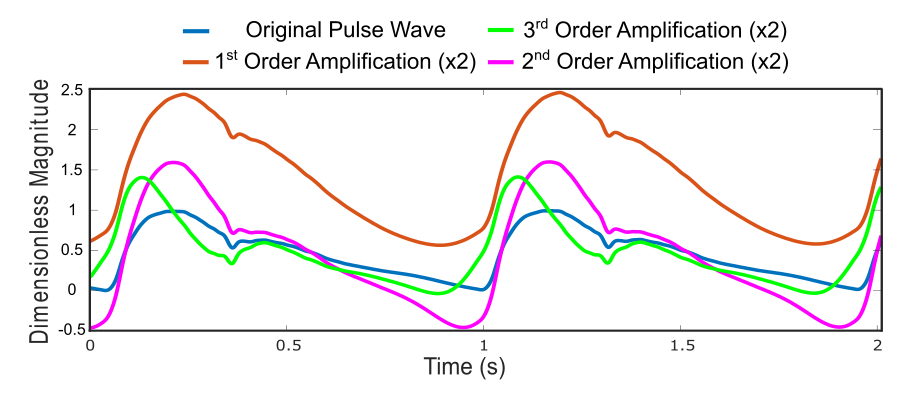}
\caption {1D distension-displacement pulse wave signal amplification, using virtual data \cite{VirtualPulseWave}. The jerk magnification shown in green creates two distinct peaks that is not present in the other two methods of lower order.}
\label{fig:OneDimensionCompare}
\end{figure} 

Fig.~\ref{fig:MagCompareBox} shows an apprehensible overview of our video magnification investigation. The pulse from the external iliac artery can be seen in the right corner and the obturator artery on the front face. Large distortion and blur can be observed on the linear magnification example, particularly in the front right corner, where as this is not present on the non-linear example, as change in velocity is exaggerated, where as any velocity is exaggerated in the linear case.
Figure~\ref{SurgicalSceneComparation} displays a magnification comparison of spatio-temporal slices taken from three different for mentioned magnification methods. E and G in this figure, demonstrates the improvement in pulse wave motion granularity using jerk has in temporal processing, compared to the lower orders. The magenta in E shows a periodic saw wave, with no discerning features relating to the underlying pulse wave signal. The magenta in G that depicts the use of acceleration shows a more bipolar triangle wave. The green in both E and G shows a consistent periodic twin peak, with the second being more diminished, which suggests that our hypothesis of a jerk temporal filter being able to detect the dicrotic notch as correct and comparable to our model analysis shown in Fig.~\ref{fig:OneDimensionCompare}.
\newcommand{\ra}[1]{\renewcommand{\arraystretch}{#1}}
\begin{table*}\centering
\ra{1.3}
\begin{tabular}{@{}rrrrcrrrcrrrcrrrcrrr@{}}\toprule
&&&
\multicolumn{3}{c}{Scene Level 1} && 
\multicolumn{3}{c}{Scene Level 2} && \multicolumn{3}{c}{Scene Level 3}  
\\ \cmidrule{4-6} \cmidrule{8-10} \cmidrule{12-14} 
$\alpha$&&Assessment &Linear&Acc.&Jerk&&Linear&Acc.&Jerk&&Linear&Acc.&Jerk\\ \midrule

$x2$ &&$\begin{array}{c}SSIM\\PSNR\end{array}$ & $\begin{array}{c}34.95\\0.94\end{array}$& $\begin{array}{c}34.7\\0.95\end{array}$& $\begin{array}{c}35.65\\0.96\end{array}$&&$\begin{array}{c}33.87\\0.93\end{array}$& $\begin{array}{c}34.33\\0.94\end{array}$& $\begin{array}{c}35.31\\0.96\end{array}$&& $\begin{array}{c}34.41\\0.95\end{array}$ & $\begin{array}{c}35.02\\0.96\end{array}$& $\begin{array}{c}35.31\\0.96\end{array}$

\\ 

$x5$ &&$\begin{array}{c}SSIM\\PSNR\end{array}$ &  $\begin{array}{c}30.6\\0.88\end{array}$& $\begin{array}{c}31.5\\0.9\end{array}$& $\begin{array}{c}33.18\\0.93\end{array}$&&$\begin{array}{c}28.98\\0.85\end{array}$ & $\begin{array}{c}31.05\\0.89\end{array}$& $\begin{array}{c}33\\0.92\end{array}$&&$\begin{array}{c}30.32\\0.9\end{array}$ & $\begin{array}{c}31.88\\0.92\end{array}$& $\begin{array}{c}33\\0.92\end{array}$  
\\
$x10$ && $\begin{array}{c}SSIM\\PSNR\end{array}$  &$\begin{array}{c}27.76\\0.82\end{array}$&$\begin{array}{c}28.94\\0.85\end{array}$& $\begin{array}{c}30.56\\0.88\end{array}$&& $\begin{array}{c}25.92\\0.78\end{array}$& $\begin{array}{c}28.41\\0.83\end{array}$ & $\begin{array}{c}30.43\\0.87\end{array}$&&$\begin{array}{c}27.75\\0.85\end{array}$ & $\begin{array}{c}29.36\\0.86\end{array}$& $\begin{array}{c}30.43\\0.88\end{array}$ 
\\
\bottomrule
\end{tabular}
\\
\caption{Results from SSIM analysis and PSNR for our surgical videos at three levels of magnification across the different temporal processing approaches.} 
\label{SSIMPSNRtable}
\end{table*}
Table~\ref{SSIMPSNRtable} shows a comparison of a surgical scene at three separate working distances. This was arranged to diminish the spatial resolution with the same objective in the endoscope.  All three aforementioned magnification algorithms were used on each at three different motion magnification ($\alpha$) factors (x2, x5, x10). 

As a comparative metric, SSIM and PSNR are used as a quantitative metric, with PSNR being based on mathematical model and SSIM taking into account characteristics of the human visual system \cite{SSIM}. SSIM and PSNR allow for objective comparisons of a processed image to a reference source, whilst it is expected that a magnified video to be altered, the residual noise generation by the process can be seen by these proposed methods. SSIM is measured in decibels (db), where the higher the number the better the quality is. PSNR is a percentile reading, with 1 being the best possible correspondence to the reference frame.
For the all surgical scene, our proposed temporal process of using jerk out performs the other low order motion magnification methods across all magnifications for SSIM and equals or outperforms the acceleration technique, particularly at $\alpha{=10}$. 
\section{Conclusion}
We have demonstrated that the use of higher order motion magnification can bring out subtle motion features that are exclusive to the pulse wave in arteries. This limits the amplification of residual signals present in surgical scenes. Our method particularly relies on the definitive cardiovascular signature characterized by the twin peaks of the end diastolic point and the dicrotic notch. Additionally, we have shown objective evidence that less noise is generated when used within laparoscopic surgery compared to other magnification technique, however, a wider sample and case specific examples would be needed to verify this claim. Further work will look at a real-time implementation of this approach as well as methods of both ground truth validation and subjective comparison within a clinical setting. Practical clinical use cases are also needed to verify the validity of using such techniques in practice and to identify the bottlenecks to translation.
\subsection*{Acknowledgements}
The work was supported by funding from the EPSRC (EP/N013220/1, EP/N027\\078/1, NS/A000027/1) and Wellcome(NS/A000050/1).


\bibliographystyle{unsrt}
\end{document}